\documentclass[letterpaper]{article}

\usepackage{natbib}
\usepackage{url}
\usepackage{hyperref}
\usepackage{amsmath}
\usepackage{cleveref}
\usepackage{booktabs}
\usepackage{amsmath}
\usepackage{graphicx}
\usepackage{subcaption}
\usepackage{caption} 
\usepackage{adjustbox}

\title{[Social] Allostasis: Or, How I Learned To Stop Worrying and Love The Noise}

\author{
    Imran Khan
    \\
    Independent Researcher \\
    imy@imytk.co.uk
} 

\begin{document}

\maketitle

\begin{abstract}
    
The notion of homeostasis typically conceptualises biological and artificial systems as maintaining stability by resisting deviations caused by environmental and social perturbations. In contrast, (social) allostasis proposes that these systems can proactively leverage these very perturbations to reconfigure their regulatory parameters in anticipation of environmental demands, aligning with von Foerster's \textit{``order through noise" }principle. This paper formulates a computational model of allostatic and social allostatic regulation that employs biophysiologically-inspired signal transducers---analogous to hormones like cortisol and oxytocin---to encode information from both the environment and social interactions, which mediate this dynamic reconfiguration. The models are tested in a small society of ``animats'' across several dynamic environments, using an agent-based model. The results show that allostatic and social allostatic regulation enable agents to leverage environmental and social ``noise'' for adaptive reconfiguration, leading to improved viability compared to purely reactive homeostatic agents. This work offers a novel computational perspective on the principles of social allostasis and their potential for designing more robust, bio-inspired, adaptive systems. 
    

\end{abstract}



\section{Introduction}

\begin{figure}[t] 
    \centering

    \begin{subfigure}[b]{0.5\textwidth} 
        \includegraphics[width=\textwidth]{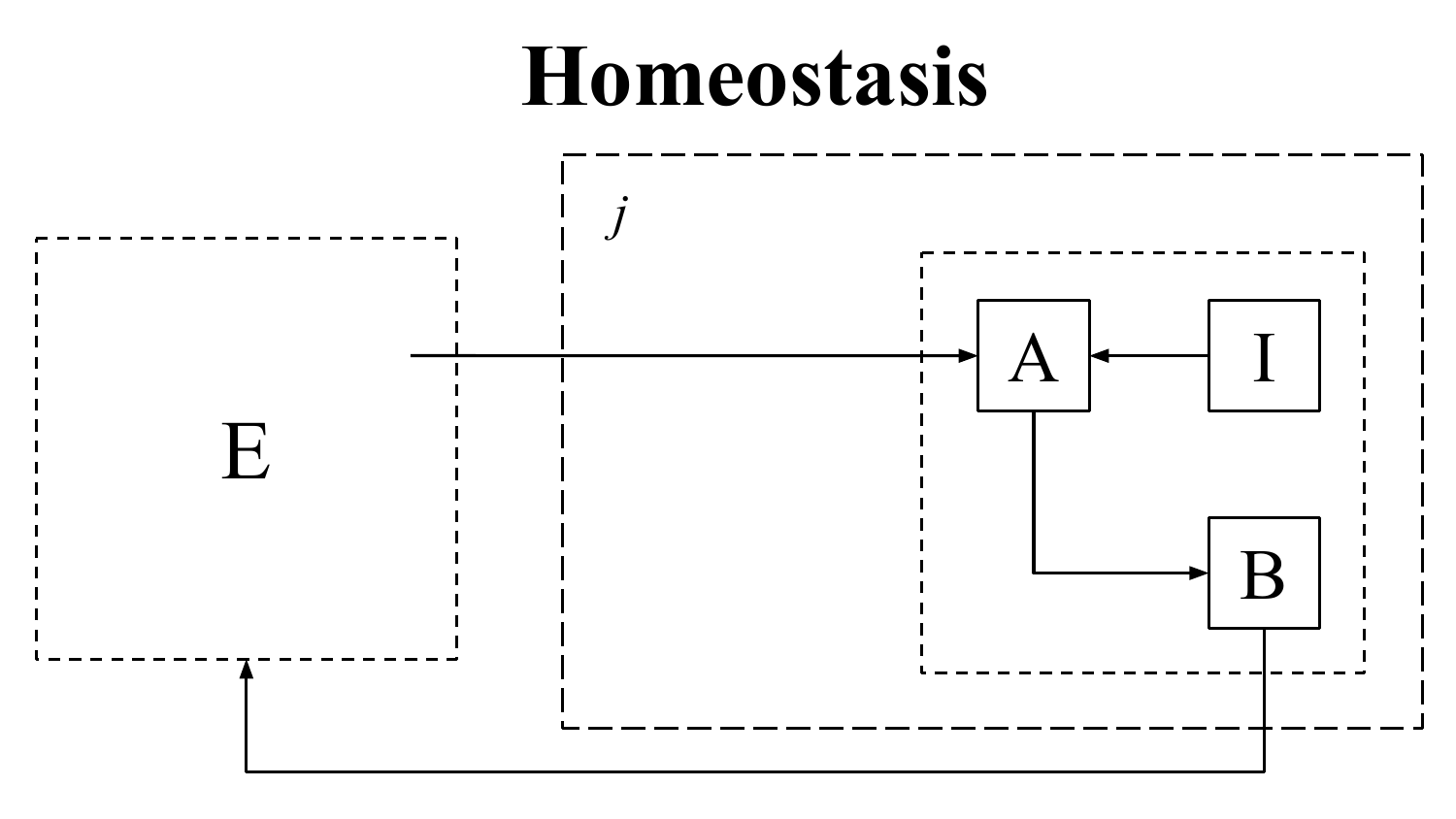}
   
        \label{fig:image1}
    \end{subfigure}
    \begin{subfigure}[b]{0.5\textwidth} 
        \includegraphics[width=\textwidth]{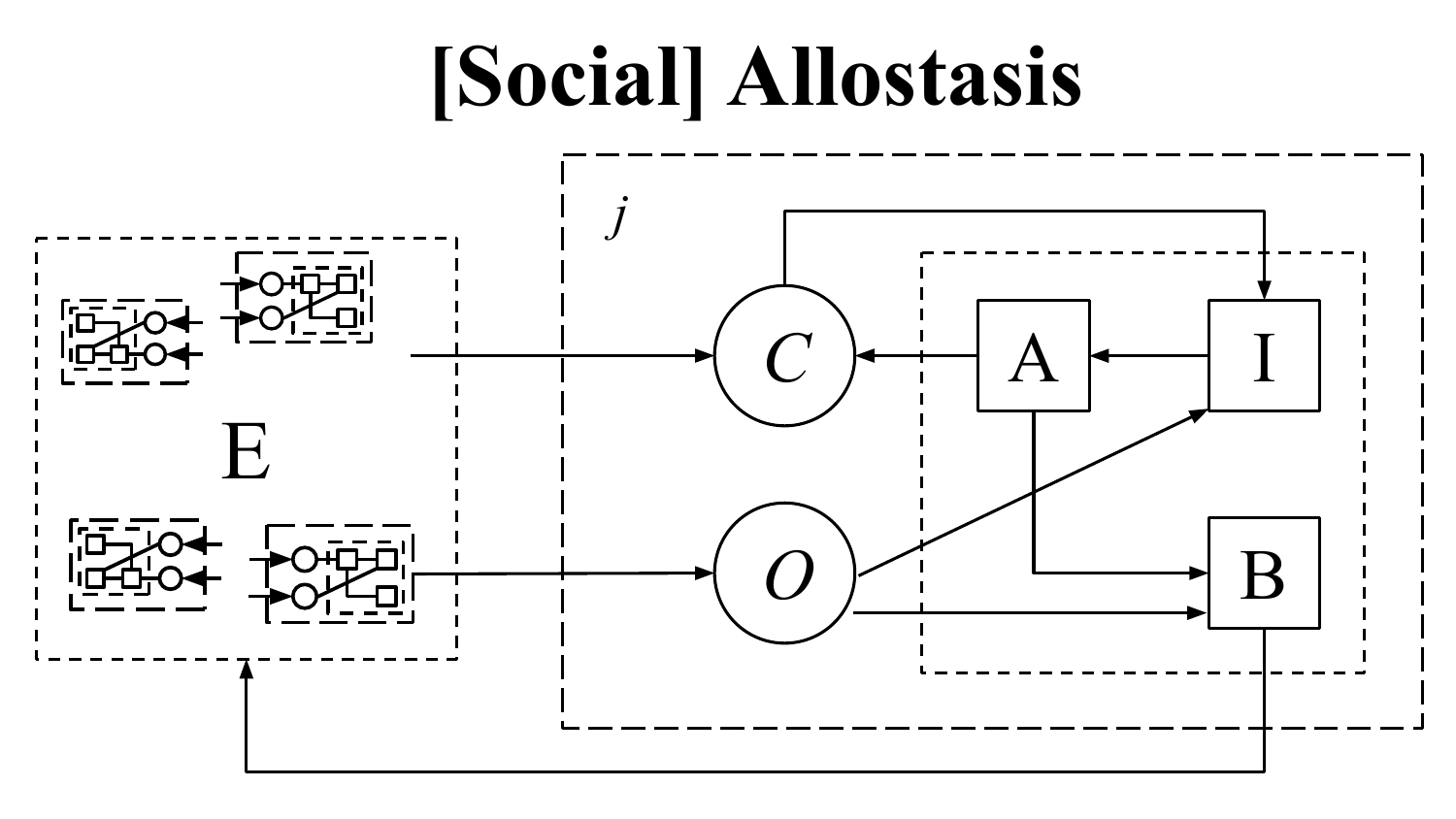}
    
        \label{fig:image2}
    \end{subfigure}

    \caption{Top: A simplified overview of homeostatic regulation. Perturbations from the environment \textit{E} influence variable \textit{A} in system \textit{j}. The difference between the regulated variable \textit{A} and its set point \textit{I} drives error-correcting behaviour \textit{B}, acting on the environment.  Bottom:  My conceptualisation of (social) allostasis discussed in this paper. External and internal information is encoded in signal transducer \textit{C}, which acts on the set point \textit{I}. In social allostatic systems, information coming from interactions with other systems is encoded in signal transducer \textit{O}, further acting on internal set points and behaviour tendencies.}
    \label{fig:imagegrid}
\end{figure}

A principal challenge for all complex adaptive systems, from individual organisms to social systems, is maintaining internal stability amidst fluctuating environments.  Classical homeostatic models emphasise rigid regulatory mechanisms that restore predetermined set points \citep{cannon_organization_1929}, an approach that has inspired much work in artificial life (e.g. \cite{montebelli_toward_2013, yoshida_homeostatic_2017, mcshaffrey_decomposing_2023}).  However, more recent conceptualisations of these mechanisms 
point to more dynamic and adaptive processes rooted in self-organisation. This process, \textit{allostasis }\citep{sterling_allostasis_2012, schulkin_social_2011}, extends homeostasis by emphasising predictive and anticipatory regulation, as  well as parameter reconfiguration, rather than strictly reactive error correction around fixed set points. 
This involves systems actively incorporating environmental information to adjust internal parameters, calibrating regulatory mechanisms to expected ecological niches and environmental demands \citep{ramsay_clarifying_2014}.  This anticipatory reconfiguration allows organisms to maintain stability \textit{through} change \citep{sterling_allostasis_2012}, rather than \textit{despite} change: a fundamental principle of adaptation across biological scales.  Social allostasis further emphasises the role of social interactions (i.e. interactions between otherwise-independent regulatory systems) and the social environment in a system's reconfiguration and stability \citep{schulkin_social_2011, saxbe_social_2020}.  Here, social dynamics are an inherent part of a system's regulatory mechanisms: maintaining stability is not solely an individual endeavour, but is deeply intertwined with (embodied) interactions with other systems. In this light, social interactions act as feedback loops, providing information that trigger allostatic responses (Figure \ref{fig:imagegrid}), enabling systems to find new stable states and behaviour possibilities.    

This dynamic regulatory process can be conceptualised through Heinz von Foerster’s ``\textit{order through noise}'' principle \citep{foerster_self-organizing_1960}. Rather than viewing perturbations as a purely disruptive force that threatens system stability (à la strictly homeostatic systems), von Foerster argued that seemingly random perturbations can instead trigger self-organising processes, allowing the system to explore its state space and find more stable attractors, resulting in improved adaptive capacity (i.e. viability) of a system.  Social allostasis provides a biologically grounded example of this principle, where the inherent \textit{noise} of the social environment---such as the unpredictability of social interactions and states of other social entities---
provides these systems with crucial information for ongoing reconfiguration of internal parameters.
Biological systems encode this environmental noise through hormonal signalling (e.g., cortisol and oxytocin), mediating these internal adjustments. Through their effects on physiological parameters (e.g., oxytocin's buffering of stress systems \citep{gunnar_social_2017})
and behavioural tendencies (e.g., approach-avoid responses \citep{cohen_oxytocin_2018}), these signals convert external variability into internal structural reconfigurations, embodying von Foerster's cybernetic principles at a biological level.

Thus, the central tenet is that the dynamic and anticipatory nature of allostatic regulation, through leveraging environmental and social perturbations as a source of information for continuous internal reconfiguration, should significantly advantage a system's long-term stability and adaptive capacity, compared to more rigid homeostatic mechanisms whose system parameters are bootstrapped to specific environmental contexts. This aligns with von Foerster's view that such pertubations can be a driving force for self-organisation.

My previous work has explored simplified models of this reconfiguration, using hormonal mechanisms to model adjustments of internal stress-related thresholds \citep{khan_adaptation-by-proxy_2021}, on social decision-making \citep{khan_long-term_2022}, and adjustments of prior preferences (i.e. homeostatic set point) within the active inference framework \citep{khan_surprise_2024}.  
Despite the theoretical benefits of (social) allostatic regulation suggested by that work, the explicit benefits over strictly homeostatic systems remain to be evaluated, particularly within social contexts and with appropriate biophysiological grounding. While some computational models have compared homeostatic and allostatic regulation in single-agent systems  (e.g. \
\citep{tschantz_simulating_2022, khan_surprise_2024}), these models are computationally intensive (requiring agents to maintain and update generative models),  lack a nuanced representation of the physiological mechanisms of allostasis, or do not evaluate these models in embodied, social contexts. 


This paper addresses these limitations by empirically evaluating the differences between strictly homeostatic, allostatic, and social allostatic regulation within a social agent-based model. Departing from purely disembodied computational approaches, I employ computationally simpler biophysiological interpretations of allostatic processes, instantiated in embodied artificial agents via agent-based modelling. I model hormonally-inspired ``signal transducers'' that encode information about the external (physical and social) environment. This paper aims to answer two questions: First, how do agents employing bio-inspired homeostatic, allostatic, and social allostatic regulatory mechanisms differ in their adaptive capacity and long-term viability across dynamic physical environmental contexts?  Secondly, to what degree does internal reconfiguration occur in these systems across these different environmental contexts?



This paper makes two primary contributions: First, it presents a novel model of social allostatic regulation that explicitly incorporates (hormonally-inspired) signal transducers to encode environmental information and influence self-organisation mechanisms in social agents.  Second, it addresses the research gap above by systematically comparing these biophysiologically-inspired allostatic and social allostatic regulatory mechanisms with traditional homeostatic approaches, evaluating their relative adaptive benefits and ability to maintain stability under environmental variability.



\section{Computational Model of (Social) Allostasis}

This section provides a description of the social allostatic agent model, with a visual overview in Figure \ref{fig:imagegrid}. To summarise, the model follows a hierarchical structure where an agent's basic physiological ``needs'' are regulated through homeostatic processes, which can be modulated by signal transducers---inspired by biological hormones, cortisol and oxytocin---that encode information about perceived environmental and social contexts. These signal transducers enable adaptive reconfiguration of regulatory parameters (i.e. (social) allostatic regulation) by acting on one or several internal set points. This architecture allows an incremental investigation into the effects of the three types of regulatory mechanisms  (homeostatic, allostatic, social allostatic). 


\subsection{Agent Model\label{subsect:agentmodel}}
Agents are artificial animals, or ``animats'', endowed with a simple, homeostatically-controlled physiology comprising two internal \textbf{drives} (or ``needs''), $\mathcal{I} = \{{\text{Energy, Socialness\}}} $. Agent action is driven by the maintenance of these two internal variables within a range of values (0--1) that allows for its ongoing viability. Each need $\mathcal{I}$  has an associated \textbf{motivational state}, $\mathcal{M} = \{{\text{Hungry, Lonely\}}} $, which, when combined with the availability of external resources $\mathcal{K} = \{{\text{Food, Agents\}}} $, results in an agent pursuing one of two consummatory behaviours $\mathcal{B} = \{{\text{Eat, Touch\}}}$. Drives, motivational states, behaviours, and resources have the following bijective mappings: $f: \mathcal{I} \rightarrow \mathcal{M}$ where $f(\text{Energy}) = \text{Hungry}$ and $f(\text{Socialness}) = \text{Lonely}$; $g: \mathcal{M} \rightarrow \mathcal{B}$ where $g(\text{Hungry}) = \text{Eat}$ and $g(\text{Lonely}) = \text{Touch}$; and $h: \mathcal{I} \rightarrow \mathcal{K}$ where $h(\text{Energy}) = \text{Food}$ and $h(\text{Socialness}) = \text{Agent}$.

This action-selection model is summarised in Equation \ref{eq:asa}. For any internal drive $i \in \mathcal{I}$ with the corresponding behaviour $b = g(f(i))$ and resource $k = h(i)$:

\begin{equation}
    b(t) = \frac{I_i - A_i(t)}{M_i} \cdot (1 + c_k(t))
    \label{eq:asa}
\end{equation}

Where $I_i$ is the homeostatic set point for drive $i$ (with defaults $I_{\text{Energy}}=0.7$ $I_{\text{Socialness}}=0.8$), $A_i(t)$ is the current value of drive $i$ at time $t$, $M_i$ is the maximum possible deficit (normalised to 1), and $c_k(t)$ is the perceived availability of the corresponding external resource $k$ at time $t$. The first factor represents the 
deficit of the internal drive (deficits), while the second factor represents how this deficit is modulated by environmental opportunities (cues). This follows the ethologically-inspired deficit-cue motivational model for multiple resource problems \citep{mcfarland_basic_1997}.

The behaviour $b \in \mathcal{B}$ with the highest intensity is selected as the winning behaviour, driving an agent either to eat a food resource or touch another agent. The successful execution of either behaviour increases the corresponding internal variable at time $t+1$ by an amount $\lambda_b$. For eating, this value is static ($\lambda_{\text{eat}} = 0.25$); for touching, this is dynamic: $\lambda_{\text{touch}} = b_{\text{touch}}(t) \cdot \kappa \cdot (1 + C(t))$ where $\kappa$ is a scaling factor, and $C(t)$ is the value of signal transducer \textit{C} at time $t$ (as calculated by Equation \ref{eq:ct_release}). In the absence of any relevant stimuli, an agent wanders randomly through the environment searching for resources, at a speed modulated by signal \textit{C}: $v_i(t) = v_0 \cdot (1 + \epsilon_t + \alpha \cdot C(t))$, where $v_0$ is the baseline speed, $\epsilon_t$ is a small random variation at time $t$, and $\alpha$ is a scaling parameter. All parameter values were established in previous work \citep{khan_modelling_2020}.

\subsubsection{Social Behaviours}
The social behaviour $b_{\text{touch}}$ is expressed either as a positive (\textit{Groom}) or a negative (\textit{Aggression}) interaction. The type of behaviour depends on the specific value assigned to each agent (see \textbf{``Partner Value Assessment''} below), and whether an agent is in a ``stressed'' state or not (see \textbf{``Stress Threshold''}). By default, all agents preferentially \textbf{groom} with the highest-value partner, as calculated by the Partner Value Assessment (Equation \ref{eq:pva}): 
\begin{equation}
    \phi_r = \arg\max_{q \in \mathcal{J}_r} V_{rq}
\end{equation}

where $V_{rq}$ is the partner value assigned to agent $q$ by agent $r$, and $\mathcal{J}_r$ is the set of agents that agent $r$ sees. When agents become \textbf{stressed} (i.e. if ${\textit{C}}  \geq \theta_s$) they may instead perform \textbf{aggression} against lower-ranked individuals, with a probability inversely proportional to the value of target partner $\phi_r$:
\begin{equation}
    P(\text{Aggression}_{r\phi_r}) = \beta \cdot (1 - V_{r\phi_r}) \quad \text{if } {\textit{C}}_r  \geq \theta_{sr}
    \label{eq:agg}
\end{equation}
where $\beta$ is a scaling factor (4) for this probability function. In other words, the smaller the partner value of target agent $\phi_r$, the greater the probability that agent $r$ will perform \textbf{aggression} toward them when it is stressed.


\subsubsection{Partner Value Assessment:} \label{subsec:pva} 
The selection of target partner for social interaction $\phi_r$ is determined by a partner value assessment function. Partner value is determined by the difference in ranks between agents, with this perceived difference modulated by the value of signal transducer \textit{O}. For agent $r$ evaluating agent $q \in \mathcal{J_r}$: 
\begin{equation}
    V_{rq} = R'_r - R'_q
    \label{eq:pva}
\end{equation}

Where $R'_r$ and $R'_q$ are the equalised ranks of agents $r$ and $q$ respectively, calculated as:
\begin{equation}
    R'_i = R_i \cdot (1 - \sigma(O_r)) + 0.5 \cdot \sigma(O_r)
    \label{eq:partnerval}
\end{equation}

\subsection{From Signals to System Adjustments: Hormone-like Transduction of Information} 
The model incorporates two key variables that I call ``signal transducers''. These signal transducers are both inspired by biological hormones—cortisol and oxytocin—and function as dynamic information encoders of environmental and social contexts. These are then translated into internal physiological signals that modulate regulatory parameters and behavioural tendencies, forming the foundation of (social) allostatic adaptation. Transducer \textit{C} serves as an embodied signal of environmental uncertainty and challenge, rising in response to homeostatic errors and resource scarcity. Transducer \textit{O} functions as an embodied signal of social support, increasing during positive interactions with other agents. 

\subsubsection{Cortisol-like Signal Transducer C}
\label{subsec:cortisol}
Transducer $C$ is a cortisol-like signal, which encodes information about environmental and physiological uncertainty. It is updated as a function of both internal homeostatic errors and external resource availability:  
\begin{equation}
\frac{dC}{dt} = \alpha \cdot (\overline{E_s} - \overline{S_k})  
\label{eq:ct_release}
\end{equation}
where $\alpha$ is a scaling parameter (0.005), $\overline{E_s}$ is the mean of the errors of internal variables (deviation from set points), and $\overline{S_k}$ is the mean of available resources: $S_{\text{food}}$ is equivalent to the amount of food an agent can see, whilst $S_{\text{agent}}$ is equivalent to the maximum partner value $V_{rq}$ that the agent can see. \textit{C} also increases in recipient agents after an \textbf{aggression} encounter, at a rate of $\eta \cdot b_{\text{touch},r}$, where $b_{\text{touch},r}$ is the intensity of the acting agent, and $\eta$ is a scaling parameter (0.3). This formulation captures how biological cortisol functions as an information carrier about multiple sources of uncertainty: internal state errors (``how far am I from homeostatic balance?'') and external resource availability (``what resources are (not) available to maintain my stability?''). This signal ranges from 0--1, with higher values representing greater environmental and physiological uncertainty.  

\subsubsection{Stress Threshold} 
Additionally, when $C$ levels exceed a threshold ($\theta_s$), the agent enters a ``\textbf{stressed}'' state characterised by increased likelihood to perform negative interactions (aggression) toward lower-ranked individuals (Equation \ref{eq:agg}), as well as increased movement speed, and heightened avoidance responses to higher-ranked individuals. 

\subsubsection{Allostatic Effect of Transducer C on Internal Set Point}
Transducer \textit{C} serves an allostatic function by acting on the set point for the \textit{Energy} variable $I_{\text{Energy}}$ as follows:
\begin{equation}
   \frac{dI_\text{Energy}(t)}{dt} = -k \cdot \frac{dC(t)}{dt} + \gamma \cdot (I_{\text{base}} - I_\text{Energy}(t))
   \label{eq:cortisol_allostasis}
\end{equation}
where $dC(t)/dt$ is the rate of change of \textit{C} at time $t$ (Equation \ref{eq:ct_release}), $-k$ is a sensitivity factor (set to 2.0), $\gamma$ is a drift rate parameter (0.001), and $I_{\text{base}}$ is the baseline set point (0.7). Therefore, the set point $I_\text{Energy}$ changes inversely with the rate of change of \textit{C}, decreasing when \textit{C} rises and increasing when it falls. When the levels of \textit{C} stabilise, the \textit{Energy} set point gradually drifts toward the baseline value (0.7). This mechanism enables agents to adaptively reconfigure their energetic needs based on environmental challenges—potentially requiring less food when resources are scarce (high stress environments) and returning to normal needs when conditions improve. The internal set point $I_\text{Energy}$ is bounded between 0.5--1.0.

\subsubsection{Oxytocin-like Signal Transducer, O}
The second transducer, $O$, is modelled after the biological hormone oxytocin, which encodes information related to an agent's social environment. In this model, \textit{O} is activated as a function of positive social interactions with agents. It is increased in both the acting agent \textit{q} and the recipient agent \textit{r} after a positive social interaction (\textit{groom}).
$$
\Delta O =
\omega_g \cdot b_{\text{touch},i}
$$
where $\omega_g$ is a weighting factor (0.2 for the acting agent, 0.1 for the recipient agent), $b_{\text{touch},r}$ is the intensity of the touch behaviour of the acting agent (Equation \ref{eq:asa}). Intuitively, $O$ can be thought of as encoding information related to an agent's (perceived) social support at a given point in time, and decays at a constant rate of 0.01 per time step and is bounded between 0 and 1.

\subsubsection{Social Allostatic Effect of Transducer O on Stress Threshold}
In line with existing evidence \cite{wu_social_2021}, transducer $O$ functions as an allostatic ``social buffering'' signal, influencing two aspects of the agent model. First, it influences the partner value assessment (Equation \ref{eq:partnerval}), equalising perceived rank differences, thereby affecting social partner selection and the perception of external stressors (Equation \ref{eq:ct_release}). Secondly, \textit{O} adjusts the stress threshold $\theta_s$, i.e. the level of \textit{C} that causes an agent to become ``stressed'':

\begin{equation}
\theta_s(t) = \theta_{\text{base}} + \theta_{\text{max}} \cdot \frac{1}{1 + e^{-\kappa(O(t) - \mu)}}
\label{eq:socialbuffering}
\end{equation}

where $\theta_{\text{base}}$ is the baseline stress threshold (0.5), $\theta_{\text{max}}$ is the maximum possible increase in threshold (0.5), $\kappa$ is the steepness parameter (15), $\mu$ is the transition point (0.65), and $O(t)$ is the level of \textit{O} at time $t$. This formulation 
models the ``social buffering'' phenomenon, potentially allowing agents who have frequent social interactions more resilient to environmental stressors.

\section{Experimental Design and Methods}

To evaluate the effects of the proposed allostatic and social allostatic models, a 3×3 experimental design is employed, using an agent-based modelling approach. I evaluated three different \textbf{Regulation Types }(homeostatic, allostatic, social allostatic) and three different \textbf{World Types} of increasing variability and physical challenge.

\subsection{Regulation Types}
All agents are initially endowed with the basic agent model described in the \textbf{\nameref{subsect:agentmodel}} subsection. The three regulatory mechanisms differ in how internal parameters are modulated by the \textit{C} and \textit{O} signal transducers. \textbf{Homeostatic} agents maintain fixed set points ($I_{\text{Energy}}=0.7$, $I_{\text{Socialness}}=0.8$) and where the stress signal transducer \textit{C} only modulates movement speed, without any reconfiguration of internal parameters. The effects of the social signal transducer \textit{O} are not included in this model. \textbf{Allostatic} agents incorporate the effects of stress signal \textit{C} on the internal set point for Energy ($I_{\text{Energy}}$) according to Equation \ref{eq:cortisol_allostasis}. In other words, allostatic agents have the potential to reconfigure their internal set point as a function of this ``cortisol-like'' signal. Again, the effects of the social signal transducer \textit{O} are not included in this model. \textbf{Social Allostatic} agents incorporate both \textit{C}'s effect on Energy set point (Equation \ref{eq:cortisol_allostasis}) as well as the social signal transducer \textit{O}'s modulatory effect on (a) stress threshold according to Equation \ref{eq:socialbuffering}, and (b) rank perception for partner evaluation as described in Equation \ref{eq:partnerval}.

\begin{figure}
    \centering
    \includegraphics[width=.8\linewidth]{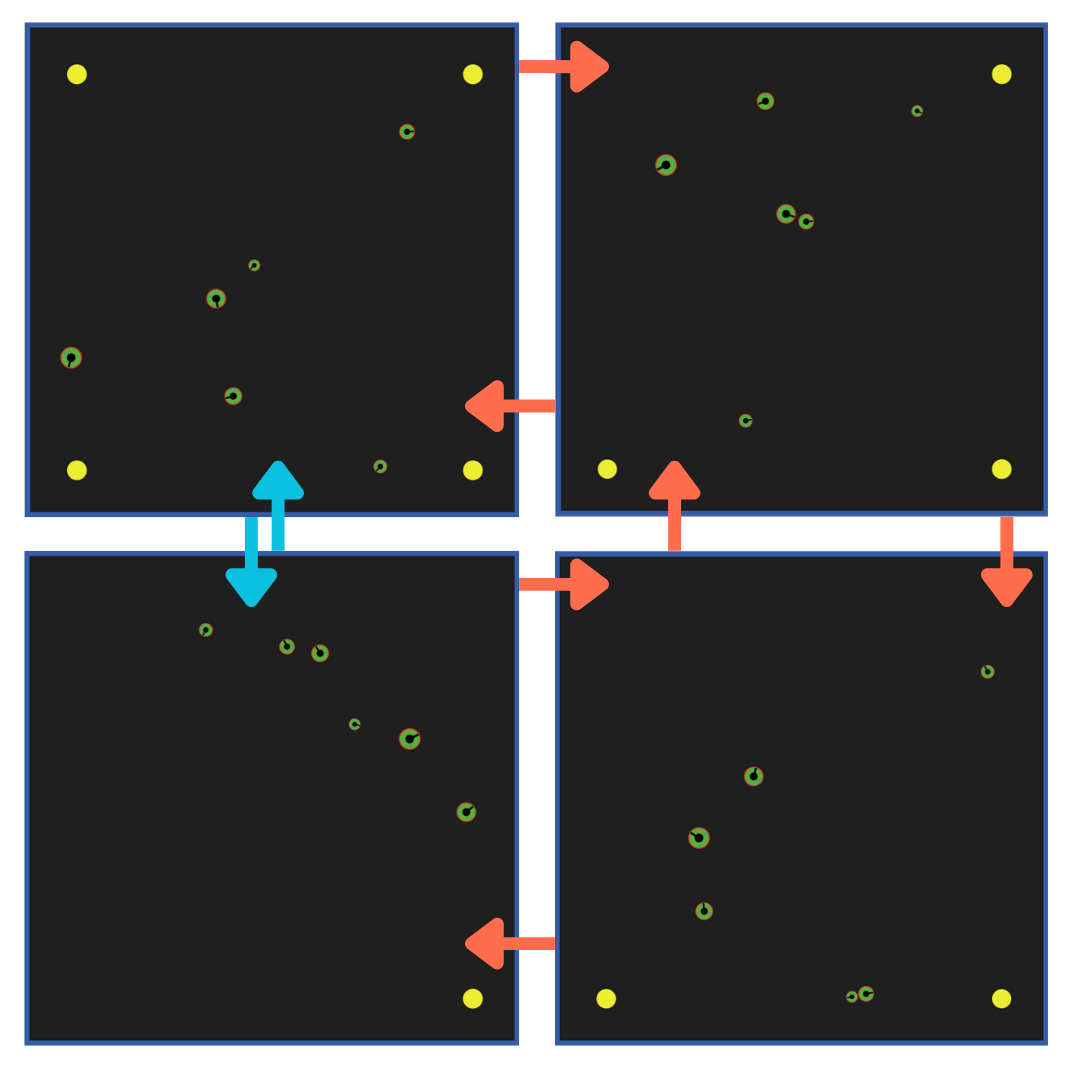}
    \caption{The different world types used in this experiment. Agents = green rings (6). Food resources = yellow circles (4). \textbf{Static}  = four food resources (top left). \textbf{Seasonal} = food incrementally changes from four to one resource and back again (orange arrows). \textbf{Extreme} = food availability changes harshly from four to one food resource (blue arrows).}
    \label{fig:environments}
\end{figure}
\subsection{World Types}
To test the adaptability of each regulatory type, I implemented three world types of increasing variability. These world types can be seen in Figure \ref{fig:environments}. 
\textbf{Static} worlds offered 
a constant of four food resources throughout the simulation and served as a control environment. \textbf{Seasonal} worlds involved gradual cycles of food availability, cycling from four down to one and back to four food items, in increments of one food resource every 2000 time steps. \textbf{Extreme} worlds involved sudden changes in food resources, switching between four or one food resource every 2000 time steps. A video of these dynamics can be found \underline{\href{https://tinyurl.com/IK-ALIFE2025}{here}}.

\subsection{Experimental Protocol}
Simulations were conducted using NetLogo v6.2.0. I initialised a society of six homogeneous agents with identical regulatory mechanisms, varying by condition. Agents differed only in their initial rank, which was randomly assigned from 1 (highest) to 6 (lowest). Simulations ran for 30,000 time steps, with 24 simulation runs per condition. Worlds were set up as a 99 $\times$ 99 grid, and all agents had a limited field of vision (of 20 units by $\pm$ 10 degrees). Initial starting positions were randomised for each run. All key model parameters (e.g., initial set points, starting values of internal variables, hormone levels) were initialised with the same values for all agents across simulation runs. 

\subsection{Data Collection and Analysis}
Several dependent variables were included as part of this experiment. To answer the first research question, I assess agent viability, defined as the total simulation time where an agent maintained their \textit{Energy} above 0; as well as physiological stability, calculated as the mean deviation of the two internal variables from their set points. 
To answer the second research question, I evaluate the overall mean changes of the internal set points ($I_\text{Energy}$ and $\theta_s$) to measure the extent of system reconfiguration across the different conditions, as well as the Shannon entropy of the former.

Data was recorded at each time step for each individual agent across all experimental conditions. Statistical analysis was performed using Python (v3.8) with the numpy, pandas, scipy, and statsmodel packages. I employed a two-way Analysis of Variance (ANOVA) to test the main and interaction effects of Regulation Type and World Type on agent viability and physiological stability, with post-hoc Tukey HSD tests for pairwise comparisons between conditions.  One-way ANOVA was used to compare the mean entropy of the internal energy set point across different regulation types, and the internal stress threshold for social allostatic agents within each world type. 

\section{Results}

\begin{figure}
    \centering
    \fbox{\includegraphics[width=\linewidth]{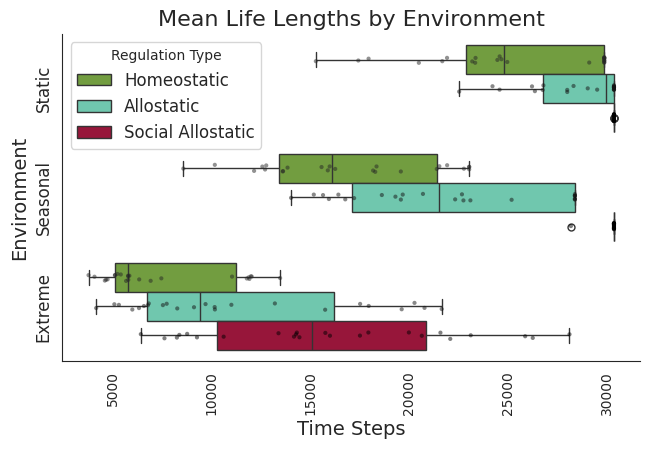}}
    \caption{Mean life length of all six agents across all regulation types, by world type.}
    \label{fig:ll_results}
\end{figure}

\subsection{Agent Viability}

\subsubsection{Life Length: }
The two-way ANOVA revealed highly significant main effects for both the regulation type (F(2,207)=86.98, p$<$0.001) and world type (F(2,207)=339.67, p$<$0.001) on mean life length. The results also found a significant interaction effect between regulation type and world type (F(4,207)=8.05, p$<$0.001), indicating that the effectiveness of the different regulatory mechanisms depends on the specific environmental conditions. Post-hoc Tukey HSD tests revealed a clear hierarchy of performance across all conditions: allostatic agents significantly outperformed homeostatic agents (0.3\%--25.95\%, M=13.15\%, p=0.04) whilst social allostatic agents demonstrated significantly greater life lengths than both allostatic agents (7.49\%--30.89\%, M=19.20\%, p$<$0.01) and homeostatic agents (18.32\%--40.3\%, M=29.31\%, p$<$0.01). Figure \ref{fig:ll_results} shows how the performance gap between regulatory mechanisms widens as environmental variability increases. This pattern strongly supports the hypothesis that higher-order regulatory processes become increasingly beneficial as environments become more challenging and unpredictable, with social allostasis offering particularly substantial advantages under conditions of high resource variability.

 
\subsubsection{Physiological Stability: } 
\begin{figure}
    \centering
    \fbox{\includegraphics[width=0.65\linewidth]{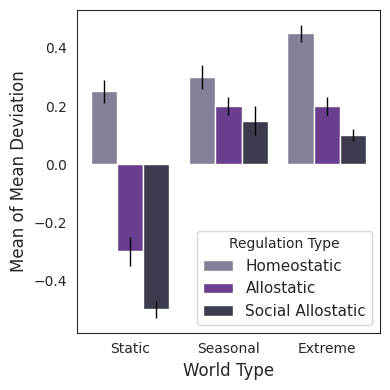}}
    \caption{Mean of mean deviations of both internal variables from internal set points (lower is better), with standard deviation bars, averaged over all agents across all regulation types per world type. }
    \label{fig:mad_results}
\end{figure}

The two-way ANOVA revealed a significant main effect of Regulation Type (F(2,18)=10.65, p=0.002) on physiological stability (MD) from internal set points. The main effect of World Type (F(2,18)=2.42, p=0.132) was not statistically significant. There was no significant interaction effect between World Type and Regulation Type (F(4,18)=1.41, p=0.288), suggesting the relative performance of regulation types was generally consistent across environments. Post-hoc Tukey HSD tests indicated that social allostatic regulation resulted in a significantly lower mean deficits compared to homeostatic regulation (M= -0.40, p=0.035). While allostatic agents showed a trend towards lower mean deficits than homeostatic agents and higher mean deficits than social allostatic agents, pairwise differences across all world types were not statistically significant. However, visual analysis and within-world comparisons suggest that the advantage of allostatic over homeostatic agents were more pronounced in the Static and Extreme worlds. Overall, social allostatic agents demonstrated the greatest physiological stability across all world conditions. This suggests that the higher-order regulatory models, and in particular those encoding social information, offer advantages in maintaining internal stability. However,  the specific benefits of allostasis may be more context-dependent on the level of environmental variability. Figure \ref{fig:mad_results} shows the mean deviation of internal values and their standard deviations across all regulation and world types.

\subsection{Agent Set Point Reconfiguration}


\begin{figure}[t] 
    \centering

    \begin{subfigure}[b]{0.9\columnwidth} 
        \fbox{\includegraphics[width=\textwidth]{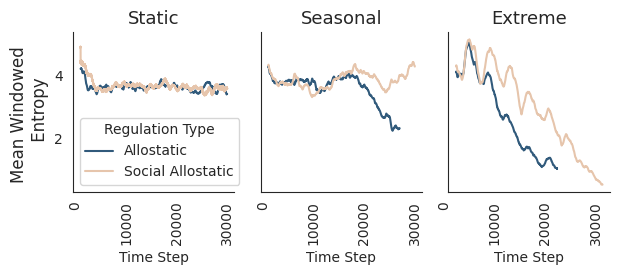}}
   
        \label{subfig:entropy}
    \end{subfigure}
    \begin{subfigure}[b]{.9\columnwidth} 
        \fbox{\includegraphics[width=\textwidth]{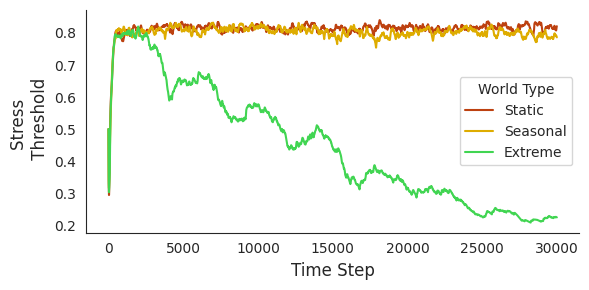}}
    
        \label{subfig:stress_threshold}
    \end{subfigure}

    \caption{Top: The mean windowed Shannon entropy of the \textit{Energy} set point ($I_\text{Energy}$). Bottom. The Stress Threshold $\theta_s$ of social allostatic agents across all world types.}
    \label{fig:dynamics}
\end{figure}

The windowed Shannon entropy of the internal set point (measured in bits) quantified the reconfiguration dynamics employed by allostatic and social allostatic agents in response to varying external environmental demands. 
This is calculated over a moving window of 1000 time steps. In 
the \textbf{Static} world type, both agent types exhibited low and similar mean entropy (allostatic: 3.64 bits, social allostatic: 3.68 bits), indicating that minimal reconfiguration was necessary for maintaining viability in these stable conditions. This aligns with earlier observations of comparable viability performance in the Static conditions, and this pattern remains relatively stable throughout the simulation.  
In the \textbf{Seasonal} world, initially (from 0--18000 time steps), both types of agents had a similar mean entropy (3.82 bits). In later stages (steps 18000-30000), allostatic agents showed a decrease to 3.08 bits, while social allostatic agents increased to 3.9 bits (p$=$0.035). Overall, allostatic agents demonstrated lower mean entropy compared to social allostatic agents (3.57 bits vs. 3.86 bits, p=0.04). The 
\textbf{Extreme} world resulted in the lowest mean entropy for both regulation types overall (allostatic: 2.83 bit, social allostatic: 3.82 bits, p$<$0.01). In the initial stages (0--5000 steps), both allostatic and social allostatic agents experienced high entropy (4.40 and 4.56 bits respectively). Later in the simulation (steps $>$5000), both agent types showed a substantial decrease. Allostatic agents decreased by 2.08 bits (from 4.4 to 2.32), while social allostatic agents decreased by 1.8 bits (from 4.76 to 2.76). While both agent types showed potential to adapt to this environment, allostatic agents exhibited a more pronounced reduction in internal adjustments, converging towards a lower-entropy strategy compared to social allostatic agents. In both the \textbf{Seasonal} and \textbf{Extreme} worlds, the higher entropy of social allostatic systems resulted in increased viability, influenced by social interactions and potentially reflecting a greater exploration of the state space and discovery of adaptive responses in the long term.

Social allostatic agents also experienced changes in the stress threshold $\theta_s$, illustrated in Figure \ref{fig:dynamics} (bottom). This internal stress threshold exhibited distinct patterns of change across environmental conditions: in the \textbf{Static} world, the mean stress threshold settled at 0.8119 (a +62.4\% change, SD = 0.056). Similarly, in the \textbf{Seasonal} world, the mean threshold reached 0.8016 (+60.3\% change from baseline, SD = 0.0614). These sustained high thresholds indicate a reduced sensitivity to minor perturbations in both the stable and slower-changing environmental contexts: agents reconfigure internal parameters to become less sensitive to small fluctuations. Conversely, the \textbf{Extreme} environment led to a significant decrease in the mean stress threshold to 0.4487 (a -10.3\% change from baseline, SD = 0.2441).
This marked reduction and high variance demonstrate a heightened sensitivity to environmental fluctuations over time, necessitating frequent and substantial adjustments to the internal stress response system in the face of unpredictability, and contributing to higher entropy in these agents.

\section{Discussion}

Overall, the results found that agents employing both types of allostatic regulatory mechanisms demonstrated statistically significant improvements in viability (i.e. life length and physiological stability) over social agents relying solely on homeostatic regulation, particularly in dynamic environmental contexts. Although comparable viability was observed between homeostatic and allostatic agents in stable environments, statistical analysis reported advantages for the latter. The results demonstrated that strictly homeostatic regulation was not sufficient for these embodied agents to overcome environmental changes in the dynamic Seasonal and Extreme worlds. Allostatic and social allostatic agents, through adjustments of internal homeostatic set points ($I_\text{Energy}$) and internal thresholds (Figure \ref{fig:dynamics}) demonstrated improved viability through a reconfiguration of internal parameters amidst environmental fluctuations. This improvement was most pronounced in the \textbf{Extreme} world, where rapid and unpredictable shifts severely challenged the capacity of homeostatic agents to maintain viability. 

Despite lacking explicit memory or cognitive models, agents with hormone-like signal transducers demonstrated a form of ``embodied memory,'' encoding environmental context: a point supported by previous findings \citep{khan_adaptation-by-proxy_2021}. Analysis of the temporal dynamics of internal parameters revealed decreasing entropy in regulatory adjustments over time, as agents discovered stable configurations that accommodated environmental fluctuations and ``predicted''  environmental demands. Unlike in homeostatic systems, where disorder was resisted, allostatic and social allostatic agents used the environmental ``noise'' as the raw material for discovering more effective regulatory states. 

\subsection{Implications for Biological Systems}

These findings suggest that the different regulatory mechanisms modelled exist on a hierarchy of increasing complexity and effectiveness: from homeostasis (defending against disturbances from fixed set points) to allostasis (using environmental uncertainty, encoded via a signal to reconfigure internal set points) to social allostasis (encoding social interactions for further reconfiguration). Each hierarchical level incorporates and builds upon the information-processing capabilities of the previous level, creating increasingly sophisticated responses to environmental challenges. 

The performance of the allostatic and social allostatic agents in the dynamic environments aligns with evolutionary evidence suggesting that more complex regulatory mechanisms, including allostasis, emerged either as adaptations to environmental unpredictability, or when organisms began to inhabit more variable niches \citep{schulkin_allostasis_2019}. 
The results from these computational models quantitatively support this hypothesis, showing a direct relationship between environmental variability and the adaptive value of allostatic regulation. Further evidence from the evolution of 
the HPA axis—the biological analogue to the stress threshold $\theta_s$—shows progressive elaboration from amphibians to mammals, with corresponding increases in regulatory flexibility \citep{denver_stress_2009}. The oxytocin system in biological systems shows evolutionary conservation, but with significant elaboration in species with complex social structures \citep{carter_oxytocin_2014}. This evolutionary trajectory mirrors the finding that (social) allostatic mechanisms demonstrate increasing advantages over homeostatic ones, particularly amidst increasing environmental and social complexities. 

\subsection{Implications for Artificial Systems}


These findings demonstrate that lightweight information encoding mechanisms can produce effective adaptive behaviours without requiring computationally expensive world models. While frameworks like active inference can endow agents with sophisticated—and often computationally expensive—generative models, the (hormone-like) signal transducers were sufficient to overcome environmental uncertainties through simple parameter modulation. These signals, therefore, offer an embodied, enactive form of world modelling, where understanding of, and adjustment to, environmental patterns emerges directly from the agents’ continuous, sensorimotor coupling with their environment. This approach aligns with Brooks' principle that ``the world is its own best model,''  \citep{brooks_intelligence_1991}  showing that minimal state variables can effectively encode and utilise relevant temporal and environmental patterns.


The model also demonstrates how social signals can provide an additional layer of environmental information without requiring explicit communication of complex world states. Instead, agents simply respond to the (perceived) success of, or interaction with others, allowing efficient information sharing through minimal signalling channels. These principles may offer a path toward creating socially adaptive artificial systems that maintain the computational efficiency of simple reactive approaches while achieving the anticipatory capabilities typically associated with more complex cognitive architectures. By implementing simple signal transduction mechanisms inspired by biological allostasis, artificial systems could develop effective (and collective) adaptive behaviours without the computational overhead of detailed world modelling.

\subsection{Implications for Cybernetics and Self-Organisation Theory}
Finally, this model contributes an instantiation of, and evidence for, von Foerster's ``order through noise'' principle, demonstrating how environmental variability drives adaptive self-organisation rather than system degradation. Analysis of the temporal dynamics of internal parameters revealed decreasing entropy in regulatory adjustments over time, as agents discovered stable configurations that accommodated environmental fluctuations. This is in line with the classic cybernetic ``law of requisite variety'' \cite{ashby_requisite_1991}: both allostatic and social allostatic agents effectively increased their internal variety via parameter reconfiguration, enabling them to match the various environmental complexities they encountered. 

This mechanism for generating requisite variety offers a computational demonstration of the ``internal model principle'', \citep{beer_brain_1972} (i.e. viable systems' regulatory components must implicitly embody models of the environmental challenges they face). The hormone-like signals, through their dynamics and internal parameter modulation, create such implicit models without requiring explicit symbolic representation.  In this way, these signals function as cybernetic ``variety amplifiers,'' enabling limited channels to convey rich environmental data via temporal integration and parameter modulation. This potentially serves as a concrete instantiation of the formal principles discussed in \cite{baltieri_bayesian_2025}, which reinterprets the internal model principle in a Bayesian context. In my model, the signal transducers and their dynamics act as a functional proxy for a Bayesian filter, where the ``belief state'' is implicitly encoded in the system's internal parameters, rather than in explicit probability distributions. These regulatory dynamics and the resulting allostatic adjustments, therefore, serve as a functional equivalent of implicit belief-updating, actively minimising the error between an agent's internal expectations (via set-point adjustment) and the environmental ``reality'' in order to maintain viability.

\subsection{Limitations and Future Directions}While my model captures key aspects of social allostatic regulation, I acknowledge several limitations. First, this implementation of hormone-like dynamics represents a simplified abstraction of their biological counterparts. This simplification, while intentional to argue for their allostatic functions via information encoding, does not capture the precise dynamics of the biological mechanisms. I also note that aggregated data at the group level does not sufficiently capture individual agent-level performance changes. Future work could explore individual agent trajectories and examine the directional information flow between regulatory systems and environmental dynamics. Formal information-theoretic measures (e.g., transfer entropy and mutual information) could quantify the statistical dependencies between individual agents, social groups, and environmental fluctuations, potentially revealing how information propagates through multi-level regulatory systems. One critical limitation is that allostatic adaptation does not come ``for free'': most interpretations of allostasis acknowledge the ``wear-and-tear'' associated with continuous adaptation. Future iterations of this model will explore this trade-off by implementing mechanisms that capture the metabolic or computational costs of maintaining flexible regulatory parameters. This will allow us to investigate how systems balance short-term adaptive capacity and long-term system longevity.

\subsection{Conclusion}

This paper has presented a computational model that demonstrates the comparative advantages of allostatic and social allostatic regulation over homeostatic approaches in dynamic physical and social environments. Homeostatic mechanisms, which resist deviations from fixed set points, exhibit limited adaptability to environmental fluctuations. Allostatic regulation improves stability by incorporating internal parameter reconfiguration, enabling agents to dynamically adjust target set points. Social allostatic regulation further enhances this capacity by incorporating social information to modulate stress thresholds and social ``perception''. This allows agents to use social interactions as an internal ``buffer'' against future environmental challenges, significantly increasing viability in dynamic environments. I demonstrate that simple signal transducers, inspired by hormones like cortisol and oxytocin, can function as information encoders that transform environmental and social variability into adaptive regulatory signals. These transducers enable a system that continuously reconfigures its internal regulatory parameters to be better suited to its ecological and social niche. Thus, allostatic and social allostatic agents harness stochasticity as the building blocks for self-organisation, providing a computational instantiation of von Foerster's \textit{order through noise} principle. 
By formalising these principles computationally, this paper offers a framework for designing artificial systems that do not merely resist environmental variability, but actively leverage it, potentially enabling more adaptive artificial social systems that, like their biological counterparts, thrive on the very uncertainty that challenges their simpler predecessors.

\footnotesize
\bibliographystyle{apalike}
\bibliography{example} 

\end{document}